\documentclass[conference]{IEEEtran}
\IEEEoverridecommandlockouts
\usepackage{cite}
\usepackage{amsmath,amssymb,amsfonts}
\usepackage{algorithmic}
\usepackage{graphicx}
\usepackage{textcomp}
\def\BibTeX{{\rm B\kern-.05em{\sc i\kern-.025em b}\kern-.08em
    T\kern-.1667em\lower.7ex\hbox{E}\kern-.125emX}}

\usepackage{times}
\usepackage{epsfig}
\usepackage{graphicx}
\usepackage{amsmath}
\usepackage{amssymb}
\usepackage{longtable}
\usepackage{supertabular}
\usepackage[table,xcdraw]{xcolor}
\usepackage{multirow}
\usepackage{amssymb}
\usepackage{pifont}

\newcommand{\xmark}{\ding{55}}
\usepackage{color}
\definecolor{light-gray}{gray}{0.85} 
\usepackage{mdframed}
\mdfdefinestyle{grayBackgroundFrame}{%
    linecolor=black,
    outerlinewidth=3pt,
    roundcorner=10pt,
    innertopmargin=3pt,
    innerbottommargin=5pt,
    innerrightmargin=5pt,
    innerleftmargin=2pt,
    backgroundcolor=light-gray}

\usepackage{stfloats}
\usepackage{adjustbox}
\IEEEoverridecommandlockouts
\usepackage{cite}
\usepackage{amsmath,amssymb,amsfonts}
\usepackage{algorithmic}
\usepackage{graphicx}
\usepackage{textcomp}
\usepackage{xcolor}
\usepackage{caption}
\usepackage{lipsum}

\def\BibTeX{{\rm B\kern-.05em{\sc i\kern-.025em b}\kern-.08em
    T\kern-.1667em\lower.7ex\hbox{E}\kern-.125emX}}
\renewcommand{\thefigure}{\arabic{figure}}

\makeatletter
\newcommand{\linebreakand}{%
  \end{@IEEEauthorhalign}
  \hfill\mbox{}\par
  \mbox{}\hfill\begin{@IEEEauthorhalign}
}
\makeatother

\begin{document}

\title{On Transfer-based Universal Attacks in Pure Black-box Setting}

\author{Mohammad A.A.K. Jalwana$^1$ Naveed Akhtar$^1$, Ajmal Mian$^1$, Nazanin Rahnavard$^2$, Mubarak Shah$^2$ \\
\small{$^1$} Department of Computer Science and Software Engineering, The University of Western Australia, Perth, WA 6009, Australia \\ 
$^2$Department of Computer Science, University of Central Florida, Orlando, FL 32816, USA}



\maketitle

\begin{abstract}
   Despite their impressive performance, deep visual models are 
   susceptible to transferable black-box adversarial attacks. Principally, these attacks craft perturbations in a target model-agnostic manner. However, surprisingly, we find that existing methods in this domain inadvertently take help from various priors that violate the black-box assumption such as the availability of the dataset used to train the target model, and the knowledge of the number of classes in the target model. Consequently, the literature fails to articulate the true potency of transferable black-box attacks. 
   We provide an empirical study of these biases and propose a framework that aids in a prior-free transparent study of this paradigm. Using our framework, we analyze the role of prior  knowledge of the target model data and number of classes in attack performance. We also provide several interesting insights based on our analysis, and demonstrate that priors cause overestimation in  transferability scores. Finally, we extend our framework to query-based attacks. This extension inspires a  novel image-blending technique to prepare data for effective surrogate model training. 
\end{abstract}

\begin{IEEEkeywords}
adversarial, universal, attack, black-box, transfer.
\end{IEEEkeywords}

\vspace{-5mm}
\section{Introduction}

Deep learning is fast closing the performance gap between human  and computer vision for a variety of tasks, including image classification \cite{archana2024deep, simonyan2014very}, object detection~\cite{redmon2017yolo9000}, \cite{ren2015faster} and semantic segmentation \cite{long2015fully}, \cite{chen2018encoder}. However, despite their impressive performance, deep  visual models are found vulnerable to adversarial attacks~\cite{akhtar2018threat}. 
These attacks are broadly categorized into white-box and black-box techniques~\cite{akhtar2018threat}. In white-box methods, the attacker has the complete knowledge of the target model, whereas black-box attacks assume \textit{no} knowledge of the target  model~\cite{akhtar2021attack}. 

Black-box attacks are further categorized as transferable (or transfer-based) attacks and query-based attacks in the literature \cite{akhtar2021attack}, \cite{akhtar2018threat, bajaj2024state}. 
The transferable attacks are supposed to operate under strict `\textit{no access  to the target model}' setting. They typically train a (range of) substitute model(s) to craft adversarial examples  with the  objective that the fooling behaviour will transfer  to the `unknown'  target model. 
Query-based algorithms relax the  `no access' rule by allowing repeated queries to the target model and the model response for each query is used to optimize the adversarial example.

\begin{table}[th!]
\footnotesize
\centering
\resizebox{0.80\columnwidth}{!}{%
\begin{tabular}{|l|l|c|c|}
\hline
\cellcolor[HTML]{EFEFEF}                                                  & \cellcolor[HTML]{EFEFEF}                                            & \multicolumn{2}{l|}{\cellcolor[HTML]{EFEFEF}Overlapping} \\ 
\multirow{-2}{*}{\cellcolor[HTML]{EFEFEF}Technique}                       & \multirow{-2}{*}{\cellcolor[HTML]{EFEFEF}Dataset}                   & Data                     & labels                    \\ \hline
L-BFGS \cite{szegedy2013intriguing}                              & \begin{tabular}[c]{@{}l@{}}MNIST \cite{lecun-mnisthandwrittendigit-2010}\\ ImageNet \cite{deng2009imagenet}\\ Youtube \cite{le2013building}\end{tabular}               & \xmark                          & \checkmark                          \\ \hline
Practical B.Box \cite{papernot2017practical}                              & \begin{tabular}[c]{@{}l@{}}MNIST \cite{lecun-mnisthandwrittendigit-2010}\\ GTSRB \cite{stallkamp2011german}\end{tabular}               & \xmark                          & \checkmark                          \\ \hline
I-FGSM \cite{kurakin2016adversarial}                                                                   & ImageNet \cite{deng2009imagenet}                                                                     & \checkmark                                & \checkmark                          \\ \hline
MI-FGSM \cite{dong2018boosting}                                                                  & ImageNet \cite{deng2009imagenet}                                                           & \checkmark                      & \checkmark                       \\ \hline
DI-FGSM  \cite{xie2019improving}                                                                 & ImageNet \cite{deng2009imagenet}                                                            & \checkmark                      & \checkmark                       \\ \hline
NI-FGSM \cite{lin2019nesterov}                                                                  & ImageNet \cite{deng2009imagenet}                                                            & \checkmark                      & \checkmark                       \\ \hline
SIM \cite{lin2019nesterov}                                                                      & ImageNet \cite{deng2009imagenet}                                                            & \checkmark                      & \checkmark                       \\ \hline
TAP \cite{zhou2018transferable}                                                                      & ImageNet \cite{deng2009imagenet}                                                            & \checkmark                      & \checkmark                       \\ \hline
SemanticAdv \cite{qiu2020semanticadv}                                                               & \begin{tabular}[c]{@{}l@{}}MSCeleb1M \cite{guo2016ms}\\ CelebA \cite{liu2018large}\end{tabular}          & \checkmark                      & \checkmark                       \\ \hline
\begin{tabular}[c]{@{}l@{}}Enhancing \\ Transferability \cite{wu2018understanding}\end{tabular}      & \begin{tabular}[c]{@{}l@{}}CIFAR-10 \cite{krizhevsky2009learning}\\ ImageNet\end{tabular}         & \checkmark                      & \checkmark                       \\ \hline
Deep Fool \cite{moosavi2016deepfool}                                                                & \begin{tabular}[c]{@{}l@{}}MNIST \cite{lecun-mnisthandwrittendigit-2010}\\ CIFAR-10 \cite{krizhevsky2009learning}\\ ImageNet \cite{deng2009imagenet}\end{tabular} & \checkmark                      & \checkmark                       \\ \hline
Curl \& Whey \cite{shi2019curls}                                                             & \begin{tabular}[c]{@{}l@{}}T-ImageNet \cite{le2015tiny}\\ ImageNet \cite{deng2009imagenet}\end{tabular}       & \checkmark                      & \checkmark                       \\ \hline
Patch Attack \cite{yang2020patchattack}                                                              & ImageNet \cite{deng2009imagenet}                                                            & \checkmark                      & \checkmark                       \\ \hline
Patchwise \cite{gao2020patch}                                                                 & ImageNet \cite{deng2009imagenet}                                                           & \checkmark                      & \checkmark                       \\ \hline
Patch-wise++ \cite{gao2020patchPlus}                                                              & ImageNet \cite{deng2009imagenet}                                                           & \checkmark                      & \checkmark                       \\ \hline
\begin{tabular}[c]{@{}l@{}}Manipulating\\ Image Attributes \cite{wei2021black}\end{tabular}   & \begin{tabular}[c]{@{}l@{}}CIFAR-10 \cite{krizhevsky2009learning} \\ ImageNet \cite{deng2009imagenet}\end{tabular}         & \checkmark                      & \checkmark                       \\ \hline
\begin{tabular}[c]{@{}l@{}}Rotation Invariant \\ Attack \cite{duan2021enhancing}\end{tabular}      & ImageNet  \cite{deng2009imagenet}                                                          & \checkmark                      & \checkmark                       \\ \hline
\begin{tabular}[c]{@{}l@{}}You see what I \\ want you to see \cite{xiao2021you}\end{tabular} & ImageNet  \cite{deng2009imagenet}                                                          & \checkmark                      & \checkmark                       \\ \hline
\end{tabular}
}
\caption{\footnotesize{Statistics of overlapped training data and classification labels between surrogate and target models for the evaluation of adversarial perturbations. Second column indicates the technique description, the third column indicates the used datasets and the last two column indicates overlap between dataset and labels.}}
\label{tab:literatureIssues}
 \vspace{-5mm}
\end{table}

In principle, transfer-based attacks neither have access to the target model nor any \textit{prior} knowledge about it. 
This includes knowledge about the training dataset and the number (and labels) of classes which can be predicted by the model. The only allowed pragmatic assumption is that the target model has the capacity to predict the label of the image under consideration. This is necessary for quantifying the attack performance. 
However, surprisingly, we find that a large body of the existing literature in transfer-based attacks violate this `pure' black-box setup. The most prominent violation comes in the form of inadvertent knowledge about the training dataset. Most existing transferable black-box methods report results for transferring attacks across the models with overlapping training datasets, see Table~\ref{tab:literatureIssues}.  This paints an incorrect picture of the potency of the attack for realistic black-box scenarios.  



For a transparent analysis of the extent of threat posed by transferable attacks, we devise a flexible black-box attack framework. Our framework allows studying the role of priors over the transferability of attacks. Specifically, we demonstrate the influence of training data and the role of the number of classes in the substitute and target models over the success of computed perturbations. Our  major contributions are summarized as:
\begin{itemize}
    \item We highlight a critical flaw in the evaluation of existing transferable attacks that violates the pure black-box assumption.
    \item We propose a framework that  allows transparent evaluation of transferable attacks while respecting the pure black-box assumption.
    \item With the proposed framework, we analyze the role of training data and number of classes in transferable attack performance. Our analysis leads to multiple interesting observations.
\end{itemize}

Besides the above major contributions, we also provide a tool to accrue data from a variety of online public sources. This tool was developed to scrap a dataset of $\approx$ 2 million images of 1000  ImageNet classes used in our extensive study. Moreover, we also introduce an image-blending technique to effectively study the query-based attack schemes as an extension of our framework. 
%
%

\vspace{-2mm}
\section{Literature Review}
\vspace{-1mm}
The literature has extensively investigated adversarial attacks along the lines of model fooling, model defense and applications of attacks beyond fooling. We first discuss key contributions along these lines before discussing the salient aspects that have largely been ignored in the evaluation of transferable black-box attacks.

\noindent\textbf{Adversarial attacks:}
Szegedy et al.~\cite{szegedy2013intriguing} observed that adding imperceptible non-random perturbations to inputs can flip model predictions. This finding had a profound impact on computer vision research,
inspiring numerous  attacks on deep visual models. 
Subsequently, Goodfellow et al. \cite{goodfellow2014explaining} proposed a single step technique called Fast Gradient Sign Method (FGSM) that crafts adversarial perturbations by  gradient ascent over the loss manifold of visual models. Later, its iterative variant, called I-FGSM was proposed by Kurakin et al.\cite{kurakin2016adversarial}. The I-FGSM  takes multiple small steps over the loss manifold to compute   perturbations. Dong et al. \cite{dong2018boosting} enhanced the I-FGSM with moment variables and showed  that their modification enhances the transferability of the resulting adversarial perturbations. Similarly, Xie et al.~\cite{xie2019feature} further improved the fooling rate of the attack by incorporating differentiable transformations that include scaling and padding to the input image. 

The above-mentioned techniques and a number of recent works 
\cite{papernot2016limitations, bajaj2024state, su2019one,moosavi2016deepfool,moosavi2017universal,carlini2017adversarial, rony2019decoupling} compute  perturbations in an algorithmic manner. There is a rich body of work \cite{poursaeed2018generative, liu2019perceptual}
 that obviates the need of hand-crafted techniques by leveraging trainable generative models to craft the  perturbations. Generally, these  models generate perturbations that have higher fooling rates, higher transferability and smaller norm. 
Moreover, these models generate perturbations considerably faster than the  algorithmic schemes. These properties make generative perturbations a more suitable tool for analyzing the adversarial vulnerabilities of models~\cite{poursaeed2018generative}. 

\noindent{\textbf{Adversarial defenses:}} The susceptibility of deep visual models to adversarial perturbations is seen as a serious threat to the practical deployment of these models in sensitive applications like disease prognostics, autonomous driving and others. This has accelerated high level of research activity towards the security of these models. A plethora of techniques~
\cite{jia2019comdefend, xie2019feature, sun2019adversarial, qiu2019adversarial, akhtar2018defense}
 have surfaced to counter the adversarial perturbations. These defense techniques commonly mitigate perturbations by altering the input to remove perturbations or modification of the network either by the addition of external modules (primarily detectors) or its robustification.\\ 
\vspace{-0.5mm}
\noindent{\textbf{Beyond adversarial perspective:}} Research from the adversarial perspective has mainly focused on  attack and defense strategies. However, there are also a few contributions that  demonstrate the utility of perturbations beyond fooling. For instance, \cite{tsipras2019robustness} and \cite{woods2019reliable} observed the manifestation of salient visual features of the target class in the perturbation signal computed by attacking robustified models. Exploring this further, \cite{jalwana2020attack} demonstrated that non-robust models also learn primitive features that can be visualized via systematic computation of perturbations. In continuation,  \cite{santurkar2019computer,jalwana2020attack} extended the utility of adversarial attacks to several image synthesis tasks that include  image generation, inpainting and interactive image manipulation.

\noindent{\textbf{Attacks evaluation:}} Adversarial attacks have confirmed the existence of a blindspot in the operational input space of deep visual models. This naturally demands a careful analysis of the strength of the attack schemes to estimate the adversarial threat  in realistic scenarios where the target model details are not available -  black-box setup. Surprisingly, most of the available literature evaluates transferable black-box algorithms with the knowledge of training dataset and the list of labels in the target attack model.  This leads to non-transparent estimation of the actual threat. This widespread practice is indicated in Table \ref{tab:literatureIssues}.

Before discussing the details of our framework (in Fig.~\ref{fig:framework}), we first provide a closer look at representation learning of visual classifiers to highlight the significance of data and labels. These details motivate the proposed framework. 

\section{Framework}
\begin{figure}[htb!]
    \centering
    \includegraphics[width=0.50\textwidth]{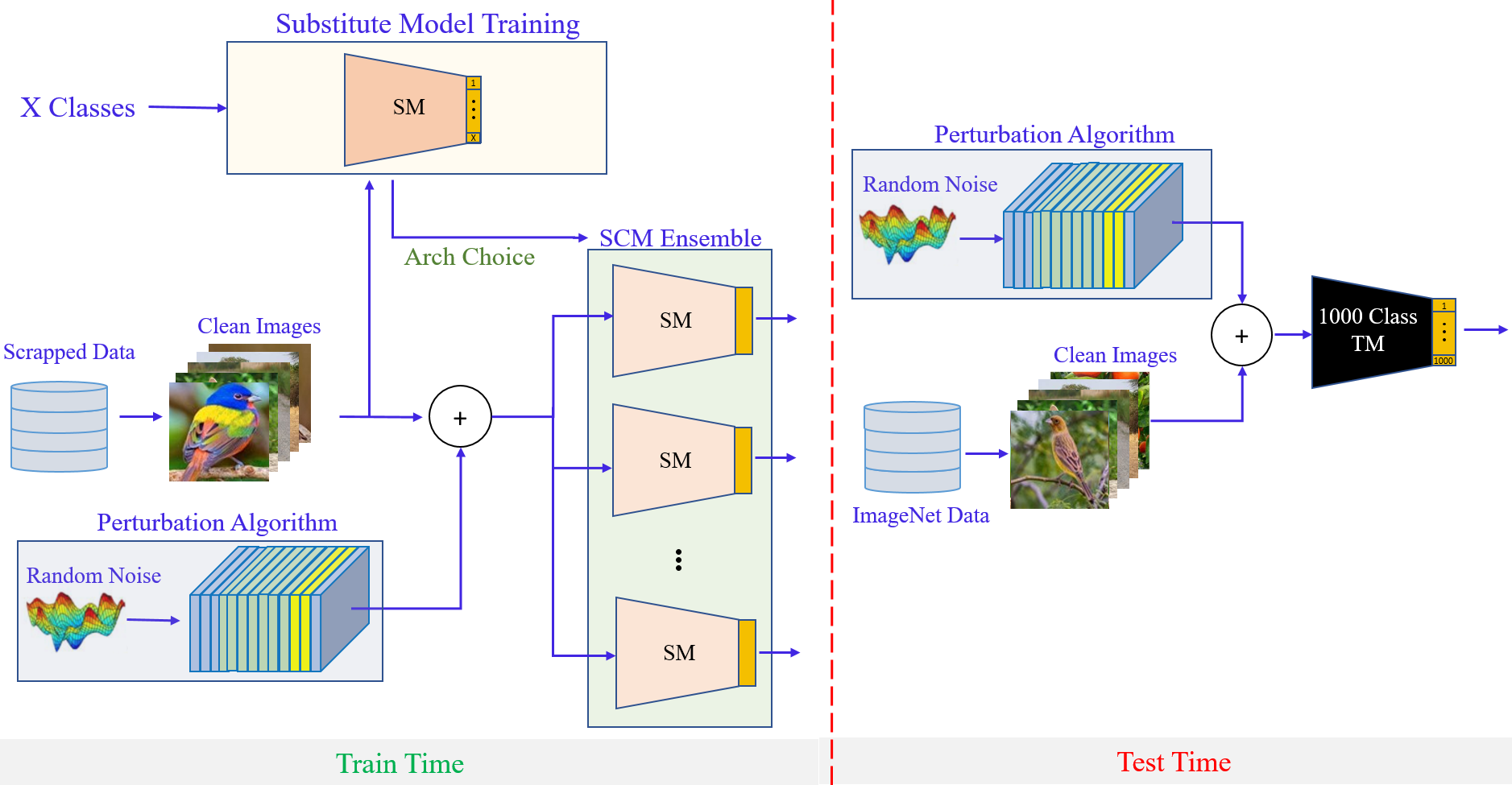}
    \caption{\footnotesize{Schematics of the proposed framework for studying the role of different priors in the transferability of adversarial perturbations. The left hand side illustrates training of  substitute models with any choice of architecture and number of classes over a scrapped training data. These models, in ensembles, in-turn train perturbation algorithms to craft perturbations for the scrapped data. The right-hand side shows testing of perturbations performed on models trained with different dataset and labels. Depiction of target model as pretrained ImageNet models is for illustration only.}}
    \label{fig:framework}
    \vspace{-3mm}
\end{figure}

\vspace{1mm}
Before discussing the details of our framework (in Fig.~\ref{fig:framework}), we first provide a closer look at representation learning of visual classifiers to highlight the significance of data and labels. These details motivate the proposed framework. 

\noindent\textbf{Representation learning:}
Fundamentally,  deep learning is a representation learning technique.  
A deep visual classifier learns 
a hierarchical
representation of concepts embedded in the input space. The objective of the learned representation is to induce a function $f(.)$  that can correctly map an input $\mathbf{x}$ to a category $\mathbf{y}$: 
\begin{equation}
     f(\mathbf{x};\theta) \rightarrow \mathbf{y} ~~\text{s.t.}~~\mathbf{x} \sim \mathrm{\mathcal{I}}(m) \in \mathbb{R}^{k},~~\mathbf{y} \in \mathbb{N}^{+}, 
\label{eq:classifierStandardEquation}
\end{equation} 
here, $\mathcal{I}(m)$ indicates the distribution of natural images for the `$m$' object categories in $k$-dimensional space, and $\theta$ denotes the parameters of the model.  Equation (\ref{eq:classifierStandardEquation}) can be casted into  an optimization problem that is solved via a variety of gradient-descent techniques \cite{simonyan2014very} .

To analyze the role of data and number of categories over the efficacy of perturbations, we provide a brief theoretical perspective over the role of distribution towards the learned representation by a classifier. 


\noindent{\textit{Lemma 3.1:}}
\textit{For $f_1(\mathbf{x}; \theta_1)$ and $f_2(\mathbf{x}; \theta_2)$ s.t.~$\theta_1 \neq \theta_2$ but $f_1: \mathbf{x} \rightarrow \mathbf{y}$ and $f_2: \mathbf{x} \rightarrow \mathbf{y}$ $\forall$ $\mathbf{x}\sim \mathcal{I}(m)$, $\rho (f_1) \approx \rho( f_2)$ where $\rho(.)$ identifies the abstract representation of $\mathcal{I}(m)$ learned by the function in its argument.}    


\noindent{\textbf{Corollary:}}~\textit{For  $f_1$(.) and $f_2$(.) trained over $\mathbf{x_1}$ and $\mathbf{x_2}$  s.t.~$\mathbf{x_1} \subset \big[ \mathbf{x} \sim \mathcal{I}(\phi) \big]$, $\mathbf{x_2} \subset \big[ \mathbf{x} \sim \mathcal{I}(\psi) \big]$ and $\phi ~\cap~ \psi \neq \emptyset$ = \{C\}  
then $f_1$(.) $\neq$ $f_2$(.) over \{C\}}.

\textit{Lemma 3.1} states that two independent models that are well-trained on the same data,  represent the same input data  distribution despite dissimilarities in their learned parameter values. Intuitively, for the same data, the differences in parameter values are compensated by other design parameters, e.g.~network architecture. This also implies that with a fixed architecture, two independently trained models (for the same data) tend to learn  similar representations. These observations explain the intrinsic transferability of attacks between different models of the same data. They also explain the amplified transferability between the models that have relatively similar architectures.  
This `representation' perspective also clarifies the observation of  higher gradient alignment between similar architectures  \cite{jalwana2020orthogonal}. 

\begin{figure}[t!]
\centering
\includegraphics[width=0.35\textwidth]{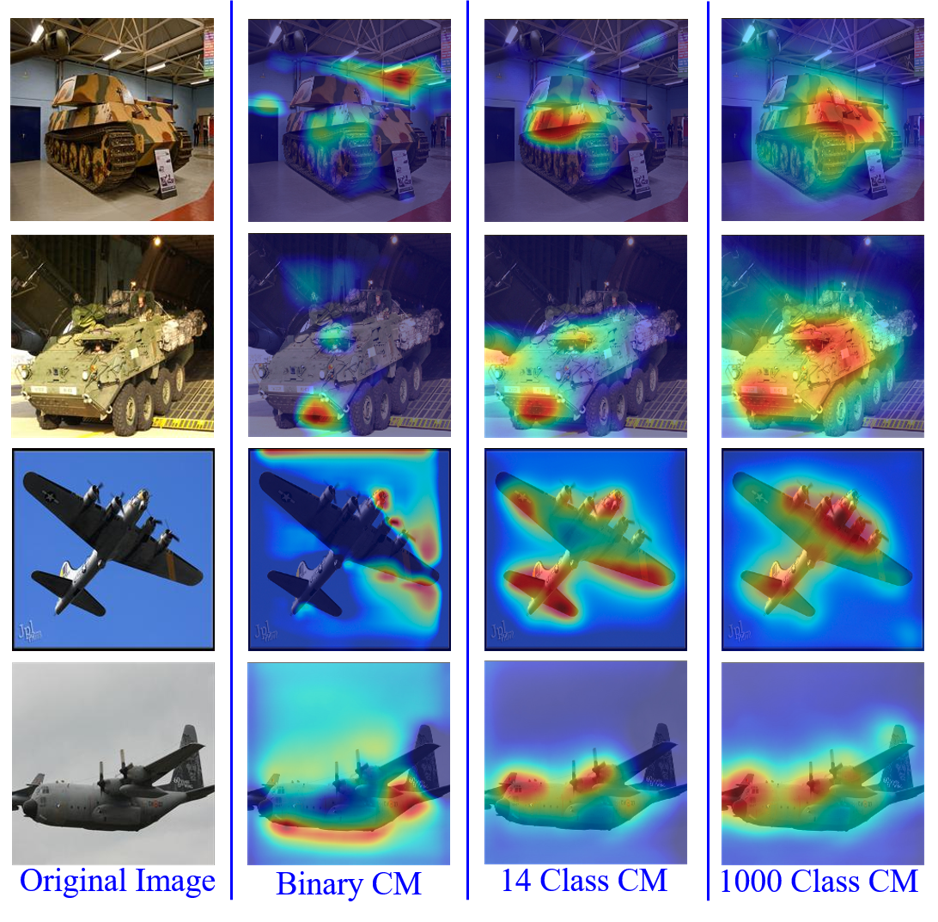}
\caption{\footnotesize{High-resolution localization of salient semantic regions with different number of classes using CAMERAS~\cite{jalwana2021cameras}. The difference in classes leads to significantly different learning of semantic concepts. Higher number of classes in  classification models  (CMs) allows better identification of the salient regions.  }}
\label{fig:cameras}
\vspace{-5mm}
\end{figure}

The \textit{Corollary} implies  that even the same network  architectures that model different distributions of data, defined by partial overlap in class labels,  can lead  to models with different learned representations. We empirically demonstrate this in Fig.~\ref{fig:cameras}. The figure  highlights the differences in model understanding for different class labels with the high-resolution and high-fidelity CAMERAS saliency maps \cite{jalwana2021cameras}. It demonstrates that besides correct prediction, these models learn  significantly different internal representations. 


For visual classifiers, the higher dimension of input space, that is typically of the order of $\mathbb{R}^{150,000}$ makes it  challenging to accrue enough training data to precisely model the actual statistical distribution of images. Therefore, the difference in training data subsets leads to modelling of different distributions, i.e.~different internal representations of models.  
Hence, it is clear that the subset of data and number of classes directly affect the data distributions modelled by the visual classifiers. This motivates us to incorporate these factors into our evaluations for a fair and transparent study of adversarial attacks.   

Motivated by the above, we  devise a framework - shown in Fig.~\ref{fig:framework} - that allows studying black-box adversarial attacks in a transparent manner.
The figure depicts a typical transfer-based attack setup that employs substitute models (SM) to compute perturbations that are later transferred to a target model (TM). 
In our analysis, we make use of scrapped data - details in the next section. 
The left hand side of Fig.~\ref{fig:framework} shows that we train a number of substitute  models over the scrapped data that allow us to subsequently compute  adversarial perturbations with our perturbation algorithm. We transfer these perturbations to a black-box target model that is trained over a `different' dataset and does not necessarily have the same number of classes or labels. 
We provide further details about the framework in the subsequent sections at appropriate locations. 

\section{Evaluation}
\label{sec:experimentation}
\subsection{Experimental setup}
In this section, we discuss the experimental setup to evaluate and analyse the role of classes and training data over the transferability of adversarial perturbations.

\noindent\textbf{Scrapped Dataset:} Availability of an independent dataset is essential to study its role in transfer-based attacks in a {\em pure} black-box setting. 
We developed a crawler in Python \cite{van1995python} that is capable of downloading images from popular online services like Flickr and Google search engine. The crawler enabled us to aggregate $\sim 2$ million images corresponding to the ImageNet class labels.  We will also publicly release the crawler and dataset for the research community. 

\noindent\textbf{Subset of Classes:}
To evaluate the impact of the number of classes in substitute models (classifiers) on the perturbation efficacy (transferability), we randomly enlisted 25 classes that most commonly appear in the outdoor from the WordNet hierarchy (nouns). These include `Trailer truck' (n04467665), `Tank' (n04389033), `Police van' (n03977966), `Military uniform' (n03763968), `Gorilla' (n02480855), `Koala' (n01882714), `Aircraft carrier' (n02687172), `Assault rifle' (n02749479), `Cannon' (n02950826), `Submarine' (n04347754), `Warplane' (n04552348), `German shepherd' (n02106662), `Ostrich' (n01518878), `Tiger cat' (n02123159), `Pet' (n01318894), `Bunting' (n01537134), `Lion cub' (n01322898), `Water polo' (n00464478), `Hydroplane racing' (n00449796), `Food' (n00021265), `Fledgling' (n01504179), `Cock' (n01514668), `Ball game' (n00471437), `Cross Country Riding' (n00451186) and `Bird' (n01503061). We chose different subsets of these classes, in ordered sequence, to train the substitute models of 2-classes, 14-classes and 25-classes. Only the first six classes are used to train perturbation generators and evaluate perturbation transferability.  

\noindent{\textbf{Data preprocessing:}} The scrapped dataset was analyzed visually to confirm that images match their label. We prepared the dataset for the above-mentioned subset of classes by manually removing the irrelevant images. This  resulted in  about $25\%$ reduction in the data. 

\noindent\textbf{Substitute Classifiers:} Our  framework trains an ensemble of substitute classifiers with varying number of classes and diverse architectures over the scrapped dataset. 
The number of classes include 2-classes, 14-classes and 25-classes and architectures include MobileNet-V3 \cite{howard2019searching},  DenseNet-121 \cite{huang2017densely}, ResNet-18/34 \cite{he2016deep} and VGG-11/16 \cite{simonyan2014very}. 

\begin{figure}[t!]
    \centering
    \includegraphics[width=0.45\textwidth]{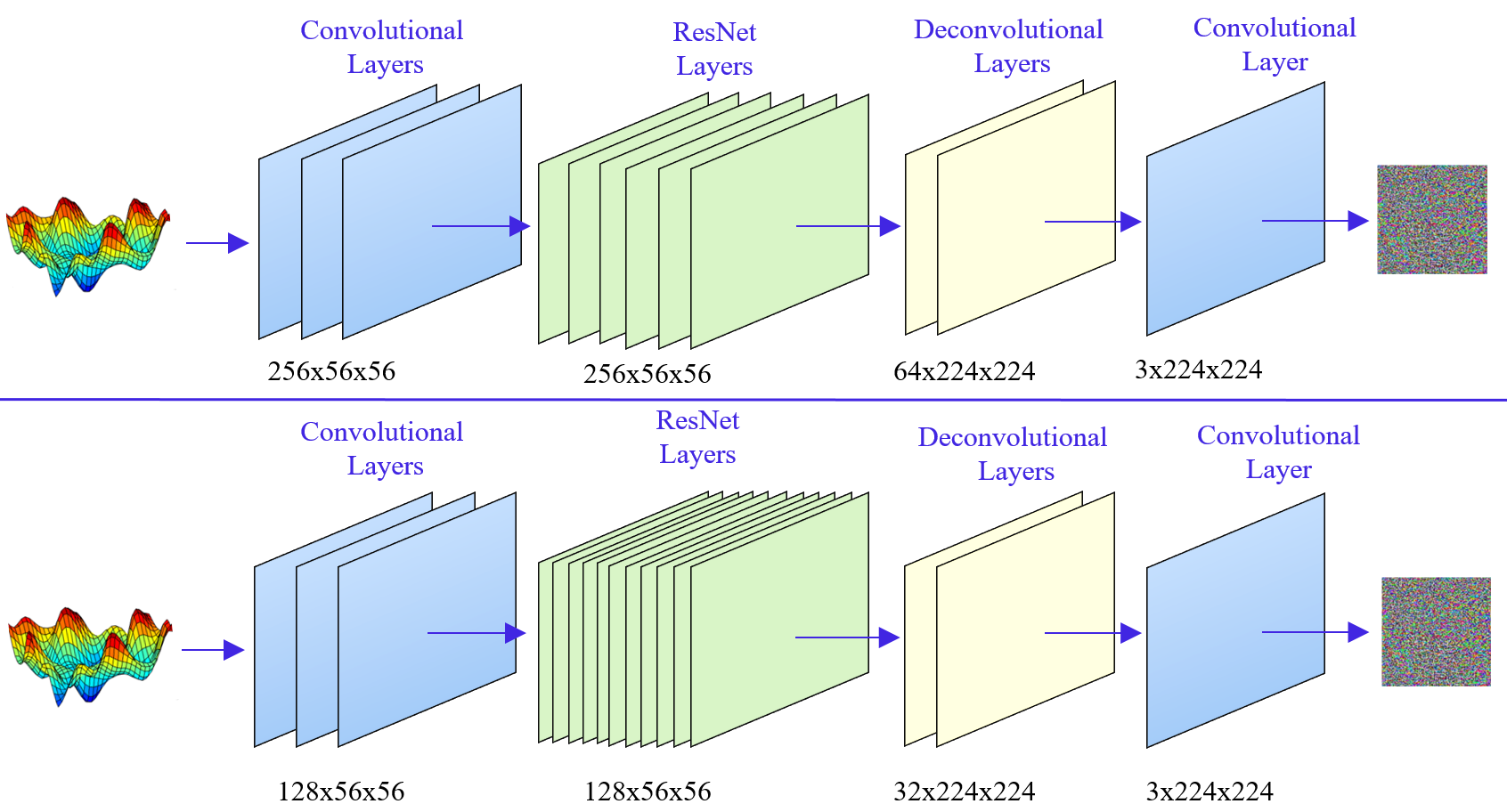}
    \caption{\footnotesize{Architectures of two variants of perturbation generators used in analysis. Top: Publicly available generator \cite{poursaeed2018generative}. Bottom: Our light-weight version.  To indicate the  differences,  output sizes of different set of layers are  also shown. }}
    \label{fig:perturbationGenerator}
    \vspace{-5mm}
\end{figure}

\noindent\textbf{Perturbation Generator:} Motivated by the previously discussed benefits of learning based perturbation algorithms, we utilized the publicly available work of Poursaeed et al.~\cite{poursaeed2018generative}. The authors devised a generative network that consists of a stack of $3$ convolutional layers, $6$ ResNet layers, followed by two de-convolational layers and a convolutional layer. This configuration is shown in of Fig.~\ref{fig:perturbationGenerator} (top). As the original configuration of \cite{poursaeed2018generative}  is proposed to learn perturbations against $1000$ class ImageNet models, we reduce the number of model parameters to create a light-weight version, which is illustrated in Fig.~\ref{fig:perturbationGenerator} (bottom). The aim of the light-weight model is to analyze if reduced parameters can aid in generating more effective perturbations. The light-weight generator has almost $50\%$ less  parameters in comparison with the original heavy-weight generator. This is mainly achieved by reducing the number of convolution filters to half for the first convolutional layer and two-fold increase in the number of ResNet layers.  

\noindent\textbf{Target Black-Box Models:} Our training loop results in a perturbation generator. 
Using that, we generate a number of perturbations and evaluate their effectiveness on the black-box models, see Fig.~\ref{fig:framework} (right). We  selected diverse popular networks  as the target black-box models,  including ResNet-50 \cite{he2016deep}, VGG-13 \cite{simonyan2014very} and MobileNet-V2 \cite{sandler2018mobilenetv2}.  

\subsection{Quantitative results}
\label{sec:QuanRes}
Our framework facilitates prior-free  transparent evaluation of black-box  attacks.  In this process, the foremost step is to train a number of substitute models that can in-turn train generators for crafting (transferable) adversarial perturbations. Finally, the crafted perturbations must be evaluated on the  black-box model to draw meaningful insights. Below, we will discuss each of these steps in detail and highlight the important observations. 

\noindent{\textbf{Substitute Model Training:}}
We train a range of substitute classification models over the scrapped data that is randomly split into training data ($\sim1000$ samples) and validation data ($\sim100$ samples) for each class. All the models are trained in Pytorch \cite{NEURIPS2019_9015} with learning rate of  $5\times10^{-4}$ and decay factor of $10$ per $30$ epochs. The stopping criterion was set by fixing the maximum number of epochs to $100$ and the model that resulted in the best validation accuracy was saved. Training statistics of the models with varying number of classes are summarized in Table \ref{tab:binarySCMs} and \ref{tab:14And25ClassSCMs}. 

\begin{table}[thb!]
\footnotesize
\centering
\resizebox{0.65\columnwidth}{!}{%
\begin{tabular}{|l|l|c|}
\hline
WordNet ID & Architecture                                                          & \begin{tabular}[c]{@{}l@{}}Validation\\ Accuracy (\%)\end{tabular} \\
\hline
n04467665  & \begin{tabular}[c]{@{}l@{}}MobileNet-V3\\ DenseNet-121\end{tabular} & \begin{tabular}[c]{@{}l@{}}97\%\\ 90\%\end{tabular}                      \\
\hline
n04389033  & \begin{tabular}[c]{@{}l@{}}MobileNet-V3\\ DenseNet-121\end{tabular} & \begin{tabular}[c]{@{}l@{}}90\%\\ 94\%\end{tabular}                      \\
\hline
n03977966  & \begin{tabular}[c]{@{}l@{}}MobileNet-V3\\ DenseNet-121\end{tabular} & \begin{tabular}[c]{@{}l@{}}91\%\\ 87\%\end{tabular}                      \\
\hline
n03763968  & \begin{tabular}[c]{@{}l@{}}MobileNet-V3\\ DenseNet-121\end{tabular} & \begin{tabular}[c]{@{}l@{}}94\%\\ 76\%\end{tabular}                      \\
\hline
n02480855  & \begin{tabular}[c]{@{}l@{}}MobileNet-V3\\ DenseNet-121\end{tabular} & \begin{tabular}[c]{@{}l@{}}99\%\\ 97\%\end{tabular}                      \\
\hline
n01882714  & \begin{tabular}[c]{@{}l@{}}MobileNet-V3\\ DenseNet-121\end{tabular} & \begin{tabular}[c]{@{}l@{}}98\%\\ 95\%\end{tabular}  \\                   
\hline
\end{tabular}
}
\caption{\footnotesize{Validation accuracy of binary  (one-vs-all) substitute models trained on the scrapped dataset. 
}}
\label{tab:binarySCMs}
\vspace{-2mm}
\end{table}

\begin{table}[]
\vspace{-3mm}
\centering
\resizebox{0.75\columnwidth}{!}{
\begin{tabular}{|l|c|c|}
\hline
Architecture & \begin{tabular}[c]{@{}l@{}}Validation\\ Accuracy (\%)\\ 14-Class\end{tabular} & \begin{tabular}[c]{@{}l@{}}Validation\\ Accuracy (\%)\\ 25-Class\end{tabular} \\ \hline
MobileNet-V3 & 82.9\%                                                                        & 78.5\%                                                                        \\ \hline
DenseNet-121 & 86.5\%                                                                        & 83.8\%                                                                        \\ \hline
VGG-11       & 83.2\%                                                                        & 79.5\%                                                                        \\ \hline
VGG-16       & 83.3\%                                                                        & 80.0\%                                                                        \\ \hline
ResNet-18    & 86.3\%                                                                        & 83.8\%                                                                        \\ \hline
ResNet-34    & 86.1\%                                                                        & 83.9\%                                                                        \\ \hline
\end{tabular}
}
\caption{\footnotesize{Validation accuracy of $14$ and $25$ class substitute models trained on the scrapped dataset. }}
\label{tab:14And25ClassSCMs}
\vspace{-1mm}
\end{table}

\begin{table}[t!]
\footnotesize
\centering
\resizebox{0.75\columnwidth}{!}{
\begin{tabular}{|l|c|c|}
\hline
Noise        & Perturbation Generator & \begin{tabular}[c]{@{}l@{}}Black-Box \\ Accuracy Drop (\%)\end{tabular} \\ \hline
Fixed        & Heavy Weight         & 10.66                                                                   \\ \hline
Distribution & Heavy Weight         & 13.88                                                                   \\ \hline
Fixed        & Light Weight         & 14.33                                                                   \\ \hline
Distribution & Light Weight         & 20.51                                                                   \\ \hline
\end{tabular}
}
\caption{\footnotesize{Average drop in the target model accuracy for attacks generated with two perturbation generator variants trained against binary substitute models. The second column indicates the type of noise used. 
The third column mentions the  generator variant. }}
\label{tab:binaryPMResults}
\vspace{-7mm}
\end{table}

\begin{table}[bp!]
\vspace{-3mm}
\resizebox{\columnwidth}{!}{
\centering
\begin{tabular}{|l|c|r|c|}
\hline
Ensemble   & \begin{tabular}[c]{@{}l@{}}No. \\ of \\ SMs\end{tabular} & \begin{tabular}[c]{@{}l@{}} No. of \\ Classes \\ in SM \end{tabular} & \begin{tabular}[c]{@{}l@{}}Black-Box \\ Accuracy \\ Drop (\%)\end{tabular} \\ \hline
MobileNet-V3 or DenseNet-121 or ResNet-18           & 1                                                        & 2                                                            & 20.51                                                                   \\ \hline
MobileNet-V3 + DenseNet-121                         & 2                                                        & 2                                                            & 22.98                                                                   \\ \hline
MobileNet-V3 + DenseNet-121 + ResNet-18             & 3                                                        & 2                                                            & 33.63                                                                   \\ \hline
MobileNet-V3 + DenseNet-121 + ResNet-18             & 3                                                        & 14                                                           & 40.33                                                                   \\ \hline
MobileNet-V3 + DenseNet-121 + ResNet-18             & 3                                                        & 24                                                           & 53.89                                                                   \\ \hline
MobileNet-V3 + DenseNet-121 + VGG-11 + VGG16        & 4                                                        & 14                                                           & 54.71                                                                   \\ \hline
MobileNet-V3 + DenseNet-121 + VGG-11 + VGG-16       & 4                                                        & 24                                                           & 57.13                                                                   \\ \hline
MobileNet-V3 + DenseNet-121 + ResNet-18 + ResNet-34 & 4                                                        & 14                                                           & 61.02                                                                   \\ \hline
MobileNet-V3 + DenseNet-121 + ResNet-18 + ResNet-34 & 4                                                        & 24                                                           & 58.11                                                                   \\ \hline
\end{tabular}
}
\caption{\footnotesize{{\bf Targeting ImageNet Models:} Results of using our substitute models and perturbation generators trained over scrapped data for targeting pre-trained ImageNet models. Average drop in target models' accuracy over the perturbations generated via both variants is reported. Each perturbation generator was trained (using the scrapped data) individually against the shown substitute model(s) in column 2. Number of substitute models (SMs) is indicated in column 3 alongside their number of classes in column 4.
This is a pure black-box setting with no overlap between the datasets used for substitute model/generator training and for the target ImageNet model training.
}}
\label{tab:classComparison}
\end{table}

The high validation accuracy in the tables indicates that our substitute models are able to generalize reasonably well.  As the number of classes increases, the average validation accuracy drops. This implies that the classifiers with larger classes must pay more attention to the discriminative  class features to maintain their performance - an observation well-supported by  Fig.~\ref{fig:cameras}. 

\begin{mdframed}[style=grayBackgroundFrame]
\footnotesize{ \textbf{Take away:} Classifiers learnt over fewer classes may resort to semantically uninteresting features for decisions.}
\end{mdframed}


\noindent{\textbf{Perturbation Generator Training:}}
We train the light-weight and heavy-weight variants of the perturbation generator over the scrapped data with the objective of fooling substitute classification models. These variants are trained with either a fixed input noise or a distribution of noise. For the fixed noise, the generator learns to transform a single input noise image to a perturbation that can fool all the training samples. For the distribution noise, the generator is trained over the signals sampled from a Gaussian distribution ($\sim \mathcal{N}(0, 0.1)$). Consequently, it learns a transformation that allows us to generate multiple perturbations. 

Table \ref{tab:binaryPMResults} summarizes the evaluation of generators trained against binary substitute models over the ImageNet data and the black-box classifiers (pretrained ImageNet) models. We can observe that fixed  perturbations do not work well against black-box models. We conjecture that a fixed perturbation encourages the generators to over-fit substitute models, thereby reducing the transferability of the perturbations. Contrarily, distributional noise encourages the generator to learn the entire manifold of perturbations  mitigating the over-fitting issue. Given the significant performance advantage ($43\%$) with the distributional noise, we limit our further analysis to the distributional noise generators.
\begin{mdframed}[style=grayBackgroundFrame]
\footnotesize{ \textbf{Take away:} Distributional noise generators are better suited to transfer-based attacks.}
\end{mdframed}

\begin{table}[htb!]
\resizebox{\columnwidth}{!}{%
\centering
\begin{tabular}{|l|c|r|c|}
\hline
Ensemble  (Pretrained-ImageNet models)    & \begin{tabular}[c]{@{}l@{}} No. \\ of \\ SMs\end{tabular} & \begin{tabular}[c]{@{}l@{}}No. of \\ Classes \\ in TM\end{tabular} & \begin{tabular}[c]{@{}l@{}}Black-Box \\ Accuracy \\ Drop (\%)\end{tabular} \\ \hline
MobileNet-V3 or DenseNet-121 or ResNet-18           & 1                                                        & 2                                                                    & 30.11                                                                   \\ \hline
MobileNet-V3 + DenseNet-121                         & 2                                                        & 2                                                                    & 29.25                                                                   \\ \hline
MobileNet-V3 + DenseNet-121 + ResNet-18             & 3                                                        & 2                                                                    & 33.79                                                                   \\ \hline
MobileNet-V3 + DenseNet-121 + ResNet-18             & 3                                                        & 14                                                                   & 43.87                                                                   \\ \hline
MobileNet-V3 + DenseNet-121 + ResNet-18             & 3                                                        & 24                                                                   & 27.31                                                                   \\ \hline
MobileNet-V3 + DenseNet-121 + VGG-11 + VGG16        & 4                                                        & 14                                                                   & 42.75                                                                   \\ \hline
MobileNet-V3 + DenseNet-121 + VGG-11 + VGG-16       & 4                                                        & 24                                                                   & 59.25                                                                   \\ \hline
MobileNet-V3 + DenseNet-121 + ResNet-18 + ResNet-34 & 4                                                        & 14                                                                   & 32.14                                                                   \\ \hline
MobileNet-V3 + DenseNet-121 + ResNet-18 + ResNet-34 & 4                                                        & 24                                                                   & 58.34                                                                   \\ \hline
\end{tabular}}
\caption{\footnotesize{{\bf Targeting our Substitute Models:} We reverse the case from Table \ref{tab:classComparison} and report results of using pre-trained ImageNet models as the  substitutes, and  target the models trained over scrapped data. Average drop in target model accuracy with both variants of perturbation generator is reported. The perturbation generators are trained individually (using ImageNet data) against the indicated substitute model(s). 
This is also pure black-box setting.}}
\label{tab:imageNetToScrappedFoolingRates}
\vspace{-3mm}
\end{table}

\noindent{\textbf{Role of classes in substitute models:}}
Our previous discussion highlights the role of the number of classes in a model's understanding of image semantics. Adversarial perturbations must corrupt semantically salient regions of images to misguide the models. Hence, the number of classes have an important link to  an attack's effectiveness. Nevertheless, this aspect has been largely ignored in the existing literature. 
Recently, many online services have surfaced that offer black-box attack evaluation~\cite{papernot2017practical}. However, besides the associated costs, their  hidden and non-flexible setup does not allow systematic and transparent study. We overcome this issue with locally trained substitute models. 

Our perturbation generators are trained over ensembles of substitute models ($2/14/25$ classes).  
Table \ref{tab:classComparison} summarizes the accuracy drop of target models by transferring these perturbations to them. 
In the table, as the number of classes in the substitute models increase, the transferability of perturbations increases correspondingly. It is important to note that $25$ class substitute models include $10$ class labels that are not present in the ImageNet target models.


\begin{mdframed}[style=grayBackgroundFrame]
\footnotesize {\textbf{Take away:} Increasing the number of classes in substitute models generally increases the transferability of adversarial perturbations. This is true even when (1) the number of classes in substitute models is still significantly lower than that of the target model and (2) the class labels do not fully overlap. }
\end{mdframed}

To further investigate, we consider the flip case where the complexity (number of classes) of substitute models is higher than the target models. This requires switching of models and datasets in the previous evaluation such that perturbation generators are trained over ImageNet validation data with pre-trained ImageNet models (as substitute models) and the perturbations are evaluated on the models trained with our scrapped dataset. Table \ref{tab:imageNetToScrappedFoolingRates} summarizes the results that reinforce the idea that as the class difference between substitute and target models decreases, we achieve higher transferability rates. 

A close inspection of the results in Table \ref{tab:imageNetToScrappedFoolingRates}  provides a valuable secondary insight about the robustness of models. We summarize this observation in Table \ref{tab:summaryOfClassRole} to show that fewer class models are more robust to adversarial perturbations than models with larger number of classes. We conjecture that this is a direct consequence of different internal representations learned by models due to different distributions of the data. The results hint towards the possible utilization of binary classifiers in practical scenarios where robustness is vital to the underlying application. Our observation also supports  recent studies \cite{jalwana2020orthogonal} that explore modification of internal representation to achieve adversarial immunity. 

\begin{mdframed}[style=grayBackgroundFrame]
\footnotesize{\textbf{Take away:} Binary classification models are inherently more robust as compared to multi-class models.} 
\end{mdframed}

\begin{figure*}[tbh!]
    \centering
    \vspace{-2mm}
    \includegraphics[width=0.75\textwidth]{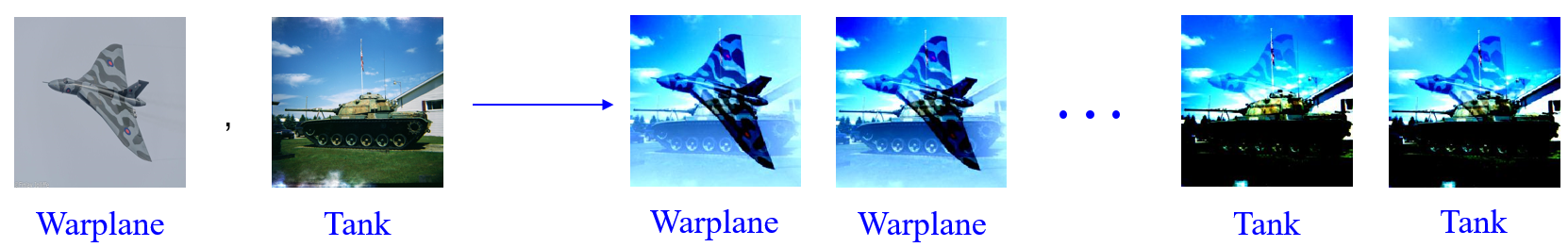}
    \caption{\footnotesize{Generating new images by linear combination of input images. The left hand-side shows sample from `warplane' and 'tank' classes that are linearly weighted and combined to generate samples on the right-hand side. }}
    \label{fig:naiveImageBlending}
\end{figure*}

\begin{table}[t!]
\centering
\resizebox{0.8\columnwidth}{!}{%
\begin{tabular}{|l|l|l|l|}
\hline
                          & \begin{tabular}[c]{@{}l@{}}Binary\\  TM\end{tabular} & \begin{tabular}[c]{@{}l@{}}14-Class\\  TM\end{tabular} & \begin{tabular}[c]{@{}l@{}}25-Class\\ TM\end{tabular} \\ \hline
Average (\%) Fooling Rate & 31.05\%                                               & 37.97\%                                                 & 49.91\%                                                \\ \hline
\end{tabular}
}
\caption{\footnotesize{Average fooling rates (\% accuracy drop) for binary, $14$-class and $25$-class substitute classification models for the perturbations generated from light-weight and heavy-weight generators. }}
\label{tab:summaryOfClassRole}
\vspace{-5mm}
\end{table}


To further highlight the role of classes, we  also consider an extreme case where the classes and training data are shared between substitute and target classification models.  Keeping the same experimental setup, the results are summarized in Table \ref{tab:imageNetToImageNetFoolingRates}. The results clearly demonstrate that priors assist significantly in achieving higher fooling rates, thus demanding prior-free evaluation of perturbations to reveal the true underlying threat posed by any scheme.

\begin{table}[h!]
\resizebox{\columnwidth}{!}{
\footnotesize
\centering
\begin{tabular}{|l|c|c|c|}
\hline
Ensemble   & \begin{tabular}[c]{@{}l@{}}No. of \\ Classes in \\ SM and TM\end{tabular} & \begin{tabular}[c]{@{}l@{}}No. \\ of \\ SMs\end{tabular} & \begin{tabular}[c]{@{}l@{}}\% Accuracy \\ drop\end{tabular} \\ \hline
MobileNet-V3 + DenseNet-121 + ResNet-18             & 1000                                                                   & 3                                                         & 57.80            \\ \hline
MobileNet-V3 + DenseNet-121 + VGG-11 + VGG-16       & 1000                                                                   & 4                                                         & 67.50             \\ \hline
MobileNet-V3 + DenseNet-121 + ResNet-18 + ResNet-34 & 1000                                                                   & 4                                                         & 71.93            \\ \hline
\end{tabular}
}
\caption{\footnotesize{Results of using pre-trained ImageNet models as both substitute and target models. Column 2 indicates the models used in the ensemble, column 3 shows that both substitute and target models have the same 1000 ImageNet classes, column 4 indicates the number of models in the ensemble and the last column summarizes the drop in accuracy of target models. }}
\label{tab:imageNetToImageNetFoolingRates}
\vspace{-3mm}
\end{table}

\begin{figure*}[b!]
\vspace{-3mm}
    \centering
    \includegraphics[width=0.75\textwidth]{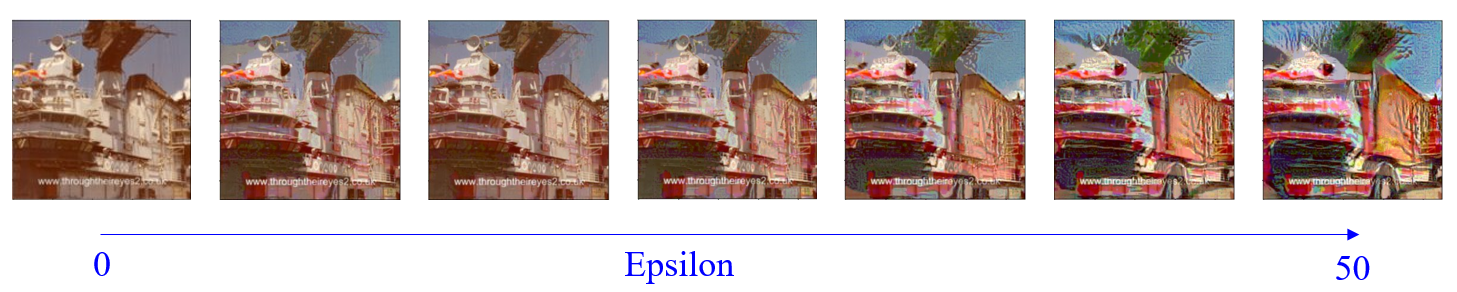}
    \caption{\footnotesize{Blending an image of `aircraft carrier' to `trailer truck' by adding perturbations computed by targeted PGD attack over robust ResNet-18 model. The left most image is the clean image, while the right image iteratively increase the norm ($\ell_{\infty}$) of perturbation gradually to a maximum value of $50/255$. }}
    \label{fig:ImageBlendingResults}
\end{figure*}

\subsection{Query Based Black-Box Attacks}

The above discussion has primarily focused on transfer based black-box perturbations that have a much stricter definition. In practical settings, it is sometimes possible to  repeatedly query the target model and access its predicted labels. The access does not reveal details of the underlying architecture or the list of labels or training data set, thereby conforming to the black-box paradigm. In the query-based attack setup, the target model can be iteratively queried to systematically align the internal representation of the substitute model with that of the target model. This can increase transferability  of perturbations crafted by the generators on the substitute model. Our framework offers a natural extension to incorporate this query-based setup and enables its transparent analysis in prior-free black-box setting. 

The existing literature highlights that a large number of queries is required to train a substitute model \cite{akhtar2021threat}. We first introduce a  technique that can generate realistic looking samples that lie in close proximity of the decision boundaries  of the target model. We then demonstrate that such images can potentially save the query volume in our query-based setup. Next, we present results demonstrating successful  attack with our extension. 

\noindent{\textbf{Image Blending:}} 
Intuitively, a substitute model can better mimic a  well-trained target model's  internal representation by correctly learning semantic concepts that differentiate between the categories, e.g.~with near-decision-boundary samples. These samples can be identified by taking a TM labeled $\ell_{o}$ image and incrementally modifying it, by some function $\beta(.)$, so that its inferred category by TM becomes $\ell_{m}\neq \ell_o$. In this iterative process, the maximally altered image that is still categorised as `$\ell_o$', and minimally altered image that is categorised as $\ell_{m}$; are two samples that necessarily lie near the decision boundary of TM along the direction of alteration (i.e.~perturbation) vector. 


A straightforward choice of  $\beta(.)$ would be an iterative linear interpolation between a given sample and a random sample from another category. Formally, this scheme can expressed as, $\beta(x_1, x_2) = \mathrm{max}(0, \mathrm{min}(\alpha x_1 + (1-\alpha)x_2, 1)) ~~\mathrm{s.t}~~  x_1, x_2 \in \mathbb{R}^{k},~\alpha \in \mathbb{R},$ and $\alpha$ is in the range $[0,1]$ sampled uniformly at random. Representative  samples generated by this approach are illustrated in Fig.~\ref{fig:naiveImageBlending}. We validate  better  alignment of SM  and TM by generation of $10^5$ samples per class under this method, and following the  experimental setup of binary classifiers with distributional-noise perturbation generator, as discussed in \S~\ref{sec:QuanRes}. Our  empirical evaluation shows a  $\sim 3.5\%$ improvement over the previous transfer-based average fooling rate.

Though effective, the above approach   suffers from two drawbacks. First,  the linear sweep over $\alpha$ intrinsically encourages large number of queries. Second, the unnatural visual quality of the generated samples has meaningless semantic information. Consequently, a large volume of naive-blended samples is needed to appropriately train the SMs. We devise a technique that addresses these issues. It generates natural looking samples by casting $\beta(.)$ as adversarial attack over a robust model i.e.~$\beta(x) = \mathrm{ProjectedGradientDescent}(\mathcal{K}(.), x, \epsilon, l)~~s.t~~ x \in \mathbb{R}^k, ~\epsilon \in \mathbb{R},~l \in \mathbb{N}^+$. Here $\mathcal{K}(.)$ is an adversarilly  robust ResNet-18 \cite{he2016deep} model that we trained over scrapped data following known adversarial training setup  \cite{madry2017towards} with $\ell_{\infty}$ budget of $15/255.0$, `$l$' is the label of target category for blending whose norm is precisely controlled by setting `$\epsilon$'. Figure \ref{fig:ImageBlendingResults} demonstrates the result of increasing norm budget over the visual quality of blended images.  

\begin{figure}[t!]
    \centering
    \includegraphics[width=0.35\textwidth]{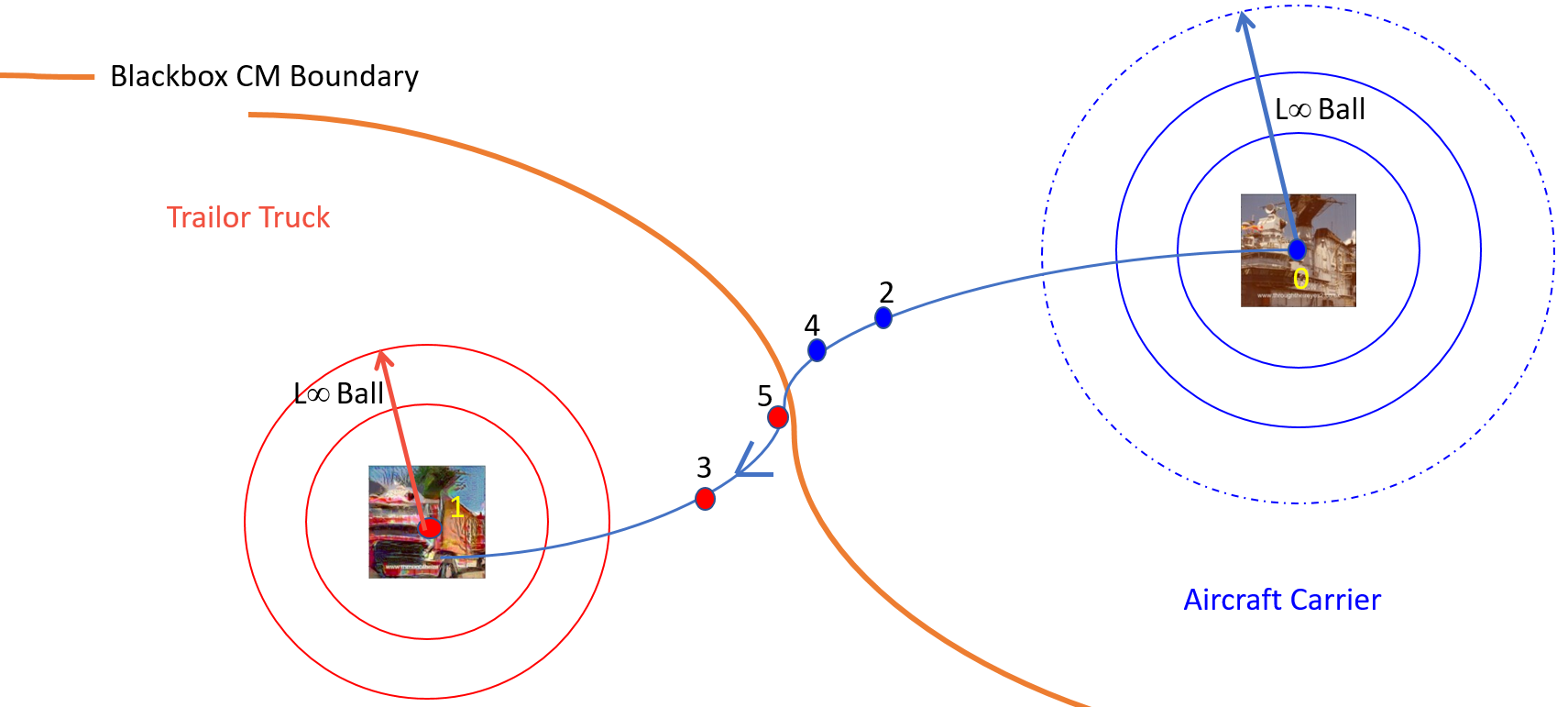}
    \caption{\footnotesize{Illustration of binary search over the blended images. A source image from `aircraft carrier' indicated as `$0$' is blended to `trailer truck' indicated as `1' by setting a large norm budget (15/255). The norm is iteratively halved to identify samples that are differently labeled by target classifier and have smaller difference in their norm, as indicated as `4' and `5'.  }}
    \label{fig:imageBlendingBoundary}
    \vspace{-5mm}
\end{figure}

Instead of linear search over the perturbation budget `$\epsilon$', we cast the problem into a binary search  to logarithmicaly reduce its computational complexity. Our  iterative process is pictorially illustrated in Figure \ref{fig:imageBlendingBoundary}. Initially, we set a  large perturbation budget, which is recursively halved to generate a blended image that is subsequently labelled by the querying TM. The near boundary samples are then identified in manner similar to the previously  discussed criterion. We set the maximum number of recursions to the statistical mode ($\approx6$) that was estimated over $1000$ sampled images blended for $25$  categories, with the aim to identify boundary samples that lie within $5\%$ norm difference.

For quantitative analysis, we repeated our experiments with binary classifiers with near boundary samples of $10^4$ samples/class from robust blending. Our evaluation identifies an overall improvement of $5.1\%$ over the transfer-based attack and a gain of $1.6\%$ over the  naive-blending with $10\times$ less training samples i.e. $10\times$ fewer queries. Thus, validating the superior performance of robust blended query-based attack. Our results verify that due to the possibility of queries, the objective of query-based attacks is relatively easier as compared to transfer-based attacks, especially when the latter is launched considering pure black-box setup.

\begin{mdframed}[style=grayBackgroundFrame]
 \footnotesize{\textbf{Take away:} Robust model-based image blending helps substitute models to mimic target model boundary better than a naive linear interpolation technique.} 
\end{mdframed}

\vspace{-3mm}
\section{Conclusion}
We made multiple contributions towards black-box adversarial attacks. First,
we  highlighted a serious flaw in the popular evaluation methodology of transferable attacks by identifying  an assumption that violates black-box setup. Second, we proposed a framework that allows transparent evaluation of attacks by avoiding any prior knowledge with respect to the black-box paradigm. With our framework, we analyzed the role of training data and the number of classes over attack performance that led to multiple interesting observations. Furthermore, we  introduced an image-blending technique to study the query-based attack schemes as an extension to our proposed framework.


\bibliographystyle{IEEEtran}
\bibliography{IEEEabrv,Utrap-Dicta-2024}

\end{document}